\documentclass[conference]{IEEEtran}
\IEEEoverridecommandlockouts
\usepackage{cite}
\usepackage{amsmath,amssymb,amsfonts}
\DeclareMathOperator{\Update}{Update}
\DeclareMathOperator{\MetaUpdate}{Meta-Update}
\usepackage{algorithm, algpseudocode}

\usepackage{graphicx}
\usepackage{textcomp}
\usepackage{xcolor}
\def\BibTeX{{\rm B\kern-.05em{\sc i\kern-.025em b}\kern-.08em
    T\kern-.1667em\lower.7ex\hbox{E}\kern-.125emX}}
\IEEEpubid{\begin{minipage}[t]{\textwidth}\ \\[10pt]
        \centering\footnotesize{\copyright 2021 IEEE.  Personal use of this material is permitted.  Permission from IEEE must be obtained for all other uses, in any current or future media, including reprinting/republishing this material for advertising or promotional purposes, creating new collective works, for resale or redistribution to servers or lists, or reuse of any copyrighted component of this work in other works.\\This paper has been accepted at DSLW 2021 for publication.}
\end{minipage}}
\begin{document}
\title{Fast On-Device Adaptation for Spiking Neural Networks via Online-Within-Online Meta-Learning}

\author{\IEEEauthorblockN{Bleema Rosenfeld}
\IEEEauthorblockA{\textit{School of Electrical and}\\\textit{Computer Engineering}\\
\textit{New Jersey Institute of Technology}\\
Newark, NJ USA }
\and
\IEEEauthorblockN{Bipin Rajendran}
\IEEEauthorblockA{\textit{Department of Engineering}\\
\textit{King's College London}\\
London, UK}
\and
\IEEEauthorblockN{Osvaldo Simeone}
\IEEEauthorblockA{\textit{KCLIP, Centre for Telecommunications Research}\\\textit{Department of Engineering}\\
\textit{King's College London}\\
London, UK}
\thanks{This work was supported by the European Research Counsil (ERC under the European Union's Horizon 2020 research and innovation programme (grant agreement No. 725731), the U.S. National Science Foundation (grants No. 1525629 and 1710009), and by Intel Labs via the Intel Neuromorphic Research Community.}}

\maketitle

\begin{abstract}
Spiking Neural Networks (SNNs) have recently gained popularity as machine learning models for on-device edge intelligence for applications such as mobile healthcare management and natural language processing due to their low power profile. In such highly personalized use cases, it is important for the model to be able to adapt to the unique features of an individual with only a minimal amount of training data. Meta-learning has been proposed as a way to train models that are geared towards quick adaptation to new tasks. The few existing meta-learning solutions for SNNs operate offline and require some form of backpropagation that is incompatible with the current neuromorphic edge-devices. In this paper, we propose an \emph{online-within-online meta-learning} rule for SNNs termed OWOML-SNN, that enables lifelong learning on a stream of tasks, and relies on \emph{local, backprop-free, nested updates}.
\end{abstract}

\section{Introduction}
\emph{Context and motivation.} The standard ``train offline-then-deploy'' approach underlies most applications of machine learning for tasks as different as facial recognition, natural language processing, and health monitoring. This framework yields rigid solutions that can produce inaccurate predictions and decisions when the data encountered after deployment presents statistical differences with respect to the training data. As an example, consider a natural language processor run by a user with a specific speech impairment or accent. Since mobile machine learning solutions are used for highly personalized tasks, there is a need to fine tune the models to the unique features of individuals and local environments after deployment in order to avoid unfair and inefficient results across different adopters of the technology \cite{kulkarni2020survey}.  

Meta-learning, or learning to learn, addresses this problem by eschewing the identification of a ``universal'' model to be fixed at deployment time, and focusing instead on determining an adaptation procedure that is able to adjust the model based on limited local data  \cite{baxter1998theoretical,stavens2007online, jiang2019improving, fallah2020personalized}. This can be generally done in one of two ways -- by meta-learning shared sub-models, e.g., feature extractors \cite{vinyals2016matching, koch2015siamese, nguyen2018meta}, or by meta-learning shared training procedures operating over local data. As an example of the latter, it is possible to meta-learn local iterative rules \cite{metz2018meta, najarro2020meta}, initializations \cite{finn2017model, nichol2018first, miconi2018differentiable}, or learning rates \cite{bohnstingl2019neuromorphic}. 
\begin{figure}[t]
    \centering
    \includegraphics[width=\linewidth, trim={1in, 4.4in, 0.7in, 2.25in}, clip]{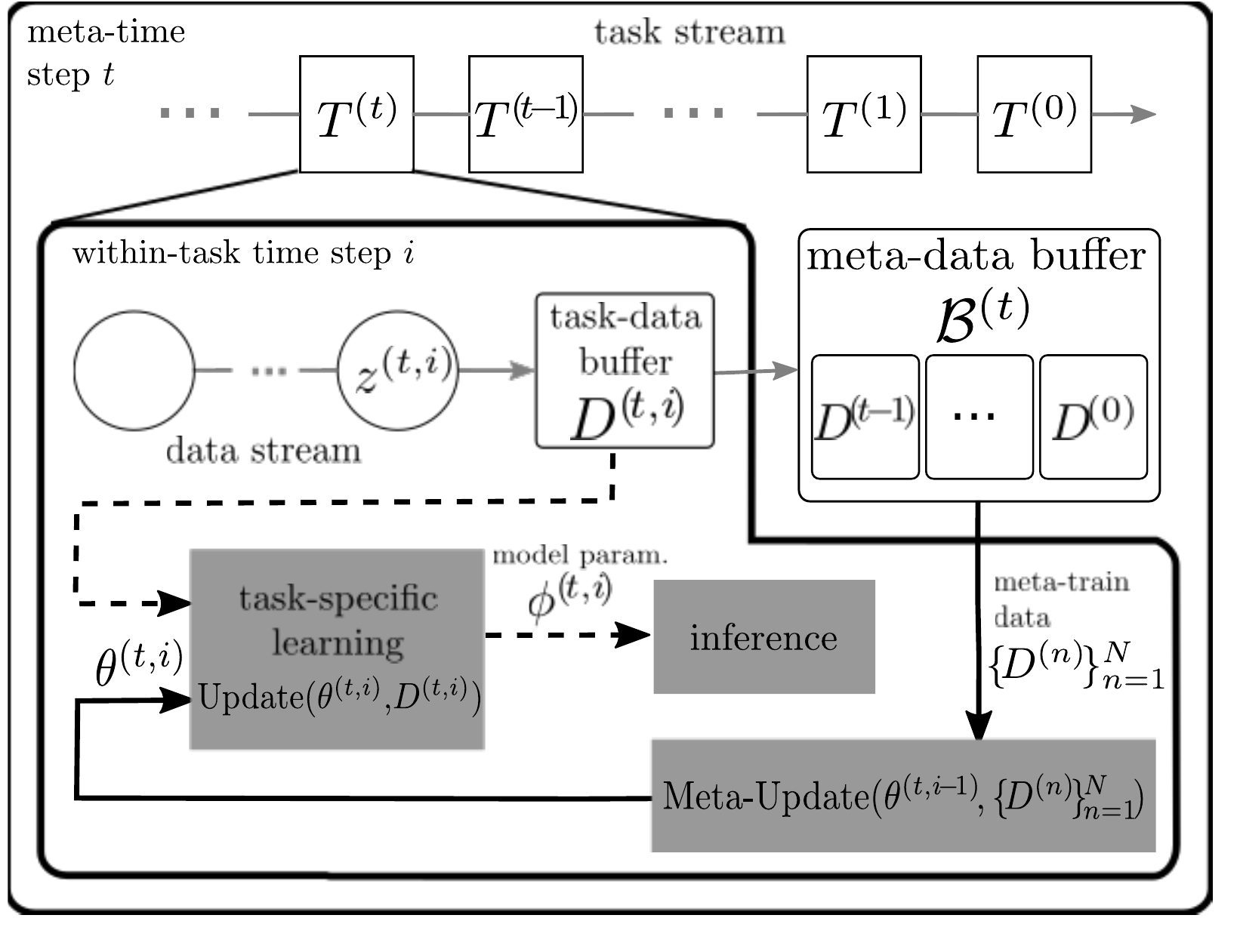}
    \caption{Online-within-online meta-learning: Tasks $T^{(t)}$ are drawn from family $\mathcal{F}$ and presented sequentially to the meta-learner over timescale $t$ (denoted in the top left corner). Within-task data are also observed sequentially (bold inset box), with a new batch $z^{(t,i)}$ added to the task-data buffer $D^{(t, i)}$ at each within task time $(t, i)$. After all within-task data has been processed, the task-data buffer is added to the meta-data buffer $\mathcal{B}^{(t)}$. At each time-step $(t, i)$ the meta-learner seeks to improve online inference by learning task-specific parameter $\phi^{(t,i)}$ using the updated task-data buffer and hyper-parameter $\theta^{(t, i)}$ as the initialization (dashed arrows). Concurrently, the meta-learner makes a meta-update to the hyperparameter, yielding the next iterate $\theta(t, i+1)$. As part of the meta-update, data from $N$ different previously seen tasks are sampled from the buffer $\mathcal{B}(t)$ and task-specific parameters for $N$ parallel models are learned starting from the hyper-parameter initialization $\theta^{(t, i)}$ (solid arrows).}
    \label{fig:owometa}
\end{figure}

\emph{Online vs offline meta-learning.} In the classical formulation of meta-learning \cite{baxter1998theoretical,jose2020information}, it is assumed that there exists a family of tasks with similarly distributed data. Tasks may, for instance, correspond to personalizations for different users. Meta-training data is obtained by sampling tasks and per-task data sets from a given, unknown, distribution. The meta-learned solution is then ``meta-tested'' on new tasks sampled from the same underlying distribution in order to evaluate its performance and adaptability. Meta-learning algorithms follow a \textit{nested loop} structure: The outer loop updates the hyperparameters that define the shared sub-models or learning procedures, and the inner loop carries out per-task learning based on limited data. 

We can distinguish different types of meta-learning problems depending on whether either loop is run using offline or online procedures. We can specifically have \emph{online/offline-within-online/offline} settings, in which the former selection applies to the outer loop and the latter to the inner loop. An offline procedure is one that uses a static data set with data points fully accessible by the algorithms, while online \mbox{(meta-)learning} uses streaming data that is sequentially processed by the algorithm \cite{denevi2019online}.

\emph{Main contributions.} The general formulation of meta-learning is in line with research in neuroscience that has shown how biological brains can learn broader concepts on a slower timescale, allowing faster adaptation to specific activities or tasks \cite{karni1998acquisition, martin2000synaptic}. Spiking Neural Networks (SNNs) are bio-inspired machine learning models that implement recurrent neural networks with sparse binary activations \cite{simeone2019learning}. With the recent development of low-power neuromorphic chips \cite{nandakumar2020experimental}, SNNs are well positioned as co-processors for on-device adaptation to new tasks. Neuron chips can implement only local learning rules whereby neurons apply updates that only depend on locally available information, with the possible use of global learning signals \cite{davies2018loihi}.

In light of this observation, this paper focuses on the development of an \emph{online-within-online} meta-learning algorithm for SNNs, termed OWOML-SNN. Unlike prior work to be reviewed in Section \ref{sec:relatedwork}, the derived rules do not require an offline pre-deployment stage and the use of backpropagation (BP). Rather, they rely on a \emph{three-factor nested local learning rule} \cite{fremaux2016neuromodulated} that is derived by following a principled approach grounded in the use of probabilistic Generalized Linear Models (GLMs) for the spiking neurons and variational inference \cite{jang2019introduction, jang2020vowel}. The online within-task updates encompass a pre-synaptic term that implements an eligibility trace on recent activity on the pre-synaptic neuron; a post-synaptic term that can be interpreted based on the principles of \emph{predictive coding} \cite{millidge2020predictive} as an error between desired or realized post-synaptic output and the probabilistic model's prediction; and a global feedback signal. The online meta-update follows the form of the first-order meta-gradient introduced in \cite{nichol2018first}.

\vspace{-0.1cm}
\section{Related Work}
\label{sec:relatedwork}

In the most common meta-learning scenario, hyperparameters are updated in the outer loop in an offline manner via a meta-gradient over data from batches of tasks selected from a data distribution \cite{gu2019meta, nguyen2018meta, finn2017model}. Batch-within-batch algorithms implement within-task learning in an offline manner as well \cite{finn2017model, nichol2018first, gu2019meta, metz2018meta}, while batch-within-online strategies use within-task data in a streaming fashion \cite{nguyen2018meta, najarro2020meta}. The online-within-online setting assumed here is most similar to that adopted in \cite{finn2019online, denevi2019online}, where an online stream of task datasets is assumed in the outer loop. In our case, as in \cite{denevi2019online}, the within-task adaptation assumes an online stream of data, while \cite{finn2019online} implements batch learning in the inner loop via standard BP. Aside from the similarity in formulation, reference \cite{denevi2019online} approaches the solution using a static deterministic linear model under a convex loss function while we study a dynamic (spiking) probabilistic model with a non-convex loss function.

All the works summarized so far assume standard ANN models. SNNs are also being explored for meta-learning due to their capability to implement within-task learning via online, local rules \cite{stewart2020online, scherr2020one, stewart2020chip, bohnstingl2019neuromorphic, bellec2018long}. In all cases, meta-training is implemented offline or pre-deployment. The SNN models adopted in prior work require the use of surrogate gradients to approximate BP, and they rely on the use of the deterministic leaky integrate and fire (LIF) neuron model. In contrast we adopt the probabilistic Generalized Linear Model (GLM) neuron model, allowing a direct derivation of local rules based on maximum likelihood \cite{jang2019introduction, jang2020vowel}.

\section{Online-Within-Online Meta-Learning}\label{sec:owoml}
In this section we outline the general operation of online within-online meta-learning via meta-gradient descent. Meta-learning assumes the presence of a family $\mathcal{F}$ of tasks that share common statistical properties. Specifically, it assumes that a common hyper-parameter $\theta$ can be identified that yields efficient learning when applied separately for each task in $\mathcal{F}$. Following the current dominant approach \cite{finn2019online,nichol2018first}, we will take hyper-parameter $\theta$ to represent the initialization to be used for the within-task training iterative procedure. 
 
In the considered online-within-online formulation adapted from \cite{finn2019online}, the meta-learner seeks to improve its \textit{online inference} capabilities over a series of tasks drawn from family $\mathcal{F}$. For each new task, it aims to quickly learn a task-specific model using streaming within-task data. To this end, the meta-learner runs an underlying meta-learning process to update the hyper-parameter $\theta$ by using data observed from previous tasks in the series. The hyper-parameter $\theta$ is then used as a within-task model initialization that enables efficient within-task training for the new task. 

To support \textit{online inference and meta-learning}, two data buffers are maintained. The \textit{task-data buffer} collects streaming within-task data used for within-task learning, While the \textit{meta-data buffer} holds data from a number of previous tasks to be used by the meta-training process. As illustrated in Fig. \ref{fig:owometa}, a stream of data sets $\mathcal{D}^{(t)}$, each corresponding to a task $T^{(t)}\in \mathcal{F}$, is presented to the meta-learner sequentially at $t=1,2,...$. Within each \textit{meta-time step} $t$, samples from data set $\mathcal{D}^{(t)}$ are also presented sequentially, so that at each \textit{within-task time step} $i$, a batch  $\mathit{z}^{(t,i)} = \{(x^j,y^j)\}_{j=1}^{B} \subseteq \mathcal{D}^{(t)}$ of $B$ training examples for task $T^{(t)}$ is observed, and added to the task-data buffer as  $D^{(t,i)} = D^{(t, i-1)} \cup \mathit{z}^{(t,i)}$ with $D^{(t,0)} = \emptyset$. Once all within-task data for task $T^{(t)}$ has been processed, the final task-data buffer $D^{(t, i)}$ is added to a meta-data buffer $\mathcal{B}^{(t)}$. 
 
The within-task training and meta-training processes take place concurrently at each time $(t, i)$. As a new batch of within-task data is observed for the current task $T^{(t)}$, the meta-learner uses it, along with the entire current task-data buffer $D^{(t,i)}$, to learn a better task-specific model parameter $\phi^{(t, i)}$ and thus improve inference on a held-out test data set. The task-specific parameter is initialized with the current hyper-parameter $\theta^{(t, i)}$ and is updated via an iterative within task training process ${\phi^{(t, i)} = \Update(\theta^{(t, i)}, D^{(t,i)})}$. Concurrently, the meta-learner improves the hyper-parameter initialization for the next round of within-task training by making a single gradient update to $\theta$. This update can be written as ${\theta^{(t,i+1)} \leftarrow \theta^{(t, i)} + \mu \nabla_{\theta}F(\theta^{(t, i)}, \mathcal{B}^{(t)})}$ for some meta-learning rate $\mu \geq 0$, where $F$ is the meta-learning objective function. The meta-learning objective evaluates the performance of the initialization $\theta^{(t, i)}$ on data from previous tasks stored in the meta-data buffer $\mathcal{B}^{(t)}$. 
\begin{figure}[t!]
    \centering
    \includegraphics[width=\linewidth, trim={1.5in, 6.85in, .5in, 2.1in}, clip]{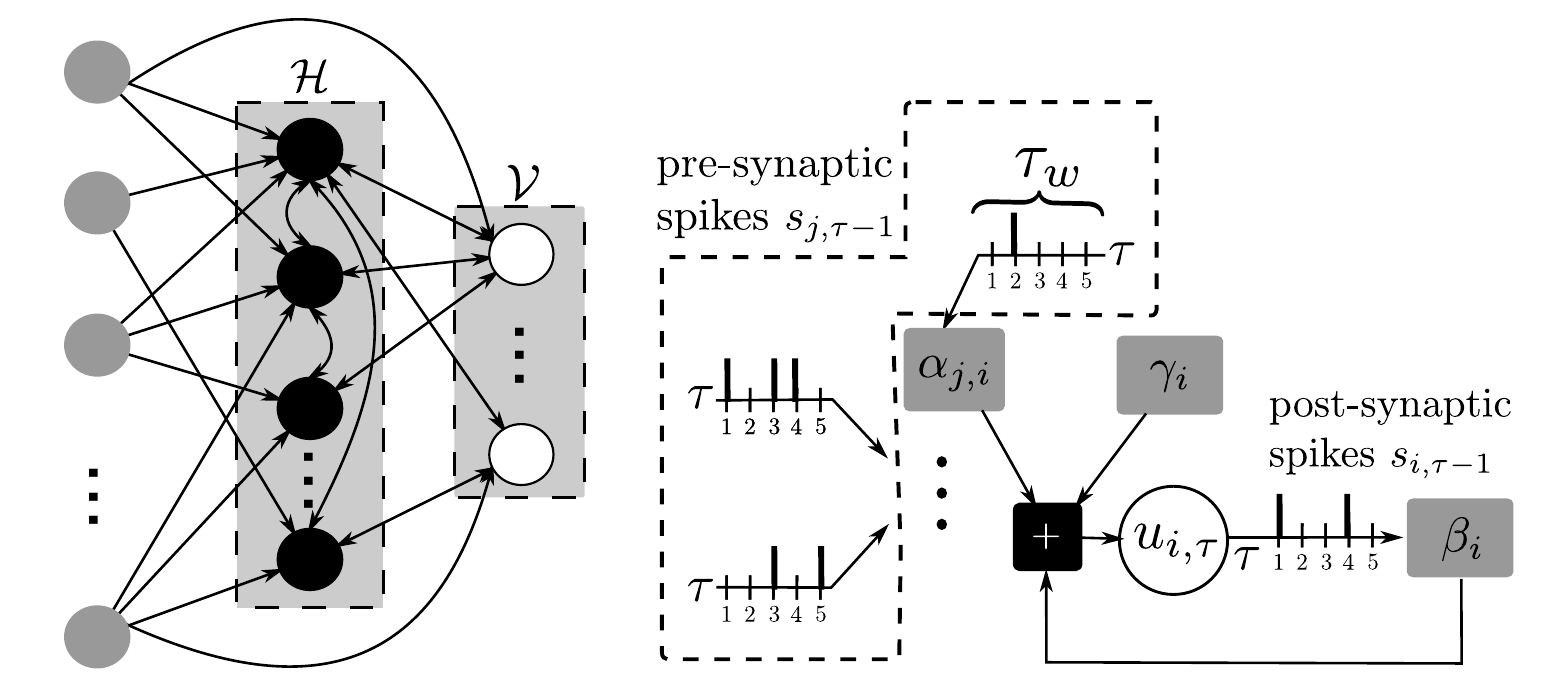}
    \caption{\emph{Left} An example of an SNN with an arbitrary topology. Black circles are the hidden neurons, in set $\mathcal{H}$, and white circles are the visible neurons in set $\mathcal{V}$, while gray circles represent exogenous inputs. Synaptic links are shown as directed arrows, with the post-synaptic spikes of the source neuron being integrated as inputs to the destination neuron. A bi-directional arrow represents two individual connections between the two neurons concerned, one in each direction. \emph{Right} A pictorial representation of a GLM neuron.}
    \label{fig:snn}
\end{figure}
Specifically, in order to evaluate the meta-learning objective function $F$, task-specific parameters for a number of previous tasks need to be learned. To this end, $N$ tasks $T^{(n)}, \ n=1,...,N,$ are drawn from the meta-data buffer $\mathcal{B}^{(t)}$ and a small data-set, $D^{(n)}$, is drawn as a subset of the stored data for each task. The within-task iterative training process is applied to learn task-specific parameters $\phi^{(n)} = \Update(\theta^{(t, i)}, D^{(n)})$ using $N$ parallel models, each initialized with the hyper-parameter $\theta^{(t, i)}$. 

The update function $\Update(\theta, D)$ addresses the problem of maximizing the likelihood of data over model parameters $\phi$. Specifically, the data set $D^{(n)}$ is split into distinct data sets $D^{(n)}_{meta}$ for the meta optimization and $D^{(n)}_{task}$  for the within-task maximization \cite{finn2017model}, and the update function tackles the problem $\max_{\phi}\log p(D^{(n)} \mid \phi)$  via stochastic gradient descent (SGD) starting from initialization $\theta$. The meta-objective is then defined as 
\begin{equation}\label{eq:meta_obj}
   F\left(\theta, \mathcal{B}^{(t)}\right) =  
   \sum_{n=1}^{N} \log p\left( 
   		D^{(n)}_{meta} \middle| \Update\left(\theta, D^{(n)}_{task}\right) 
   \right).
\end{equation}

The meta-gradient $\nabla_{\theta}F (\theta, \mathcal{B}^{(t)})$ requires the computation of the second order gradient of the training losses used in the task-specific learning function $\Update(\theta, D^{(n)}_{task})$ \cite{finn2017model}. In this work, we make use of the first-order REPTILE approximation for the gradient 
\begin{equation}\label{eq:reptile_udpate}
    \nabla_{\theta}\log p\left( D^{(n)}_{meta} \middle| \phi^{(n)}\right) \approx
    \theta^{(t, i)} - \phi^{(n)},
\end{equation}
which has been shown to have properties similar to the true gradient in a number of benchmark tasks \cite{nichol2018first}. This yields the meta-update function 
\begin{align}
    \MetaUpdate&(\theta^{(t,i)}, \{D^{(n)}\}_{n=1}^N) =\\ &\theta^{(t, i)} \!+ \mu \!\sum_{n=1}^N \!\left(\theta^{(t, i)} \!- \Update(\theta^{(t, i)}, D^{(n)}_{task})\right),\nonumber
\end{align} where $\mu \geq 0$ is the meta-learning rate.

\section{SNN Model}\label{sec:snnmodel}
In this paper, we train a recurrent SNN based on the probabilistic generalized linear model (GLM) spiking neurons  \cite{jang2019introduction} by using the online-within-online meta-learning framework described above. The SNN processes data in the form of binary signals (spikes) over SNN processing time $\tau=1,2,...$, with each neuron $i$ producing an output spike, $s_{i,\tau}=1$, or no output spike, $s_{i,\tau}=0$, at any time $\tau$. The neurons are classified into the set of visible neurons $\mathcal{V}$  and the set of hidden neurons $\mathcal{H}$, with respective spiking outputs denoted as $s_{i, \tau} = \upsilon_{i, \tau}, \ i \in \mathcal{V}$ and $s_{i, \tau} = h_{i, \tau},  \ i \in \mathcal{H}$. The spiking signals for visible neurons are specified by data during training, While the outputs of the hidden neurons are auxiliary to the operation of the visible neurons. Arbitrary connections are defined between neurons within each set and across sets, such that the post-synaptic spike of any neuron may be a pre-synaptic input to any other neuron in the network (see Fig. \ref{fig:snn}). 

Pre-synaptic and feedback spikes are integrated via time domain filtering to update the membrane potential at every time $\tau$, giving the instantaneous membrane potential for neuron $i$ 
\begin{equation}\label{eq:glm_mempot}
    u_{i,\tau} = \sum_{j} \alpha_{j,i}*s_{j,\tau-1} + \beta_i*{s}_{i,\tau-1} + \gamma_i,
\end{equation} where $\alpha$ and $\beta$ are trainable pre-synaptic and feedback filters, respectively, and $\gamma_i$ is a trainable bias. Each filter $\alpha_{j,i}$ that connects neurons $j$ and $i$ operates over a window of $\tau_w$ previous spikes. The filter is defined as a linear combination of a set of $K_a$ raised cosine basis functions collected as columns of matrix $A$, such that we have ${\alpha_{j,i} = A w^{\alpha}_{j,i}}$ where $w^{\alpha}_{j,i}$ is a $K_a \times 1$ vector of trainable synaptic weights \cite{pillow2008spatio}. The post-synaptic feedback filter $\beta_{i} = B w^{\beta}_i$ is defined similarly. We collect in vector $\phi=\{w^{\alpha}, w^{\beta}, \gamma \}$ all the model parameters.

In the GLM neuron, a post-synaptic sample $s_{i,\tau}$ is a random variable whose probability is dependent on the spikes previously integrated by that neuron, $s_{\leq \tau-1} = \{h_{\leq \tau-1}, \upsilon_{\leq \tau -1}\}$, including pre-synaptic hidden and observed spikes. It is defined as a probabilistic function of the neuron's membrane potential $u_{i,\tau}$ at that time as
\begin{equation}\label{eq:glm_spikeprob}
     p_{\phi}(s_{i,\tau} \mid s_{\leq \tau -1}) = p_{\phi}(s_{i,\tau}=1\mid u_{i, \tau}) = \sigma(u_{i,\tau}),  
\end{equation} where we have $\sigma(x) = (1+e^{-x})^{-1}$ and $\phi$ is the vector of trainable model parameters. 

The likelihood of any sequence of observed spikes $\upsilon_{\leq \mathcal{T}} = \{[\upsilon_{i, 0},...,\upsilon_{i, \tau},...,\upsilon_{i, \mathcal{T}}]\}_{i \in \mathcal{V}}$ conditioned on some sequence of hidden spikes $h_{\leq \mathcal{T}-1}=\{[h_{i,0},..., h_{i, \tau},..., h_{i,\mathcal{T}-1}]\}_{i\in\mathcal{H}}$ can then be defined as
\begin{align} \label{eq:glm_outprob}
    p_{\phi} (\upsilon_{\leq \mathcal{T}} \mid \mid h_{\leq \mathcal{T}-1}) = \prod_{\tau=1}^{\mathcal{T}} \prod_{i\in\mathcal{V}} (&\overline{\upsilon}_{i,\tau} \overline{p}_{\phi}(\upsilon_{i, \tau} \mid u_{i,\tau})\nonumber
    \\& + \upsilon_{i,\tau}p_{\phi}(\upsilon_{i, \tau} \mid u_{i,\tau})),
\end{align}
where we have used the causally conditioned notation \cite{kramer1998causal} ${p(a_{\leq \mathcal{T}} \mid \mid b_{\leq \mathcal{T}-1}) = \prod_{\tau=1}^{\mathcal{T}}p(a_{\tau} \mid a_{\leq \tau - 1}, b_{\leq \tau-1})}$ as well as the notation $\overline{x} = (1-x)$. Similarly, the membrane potential of the hidden neurons defines the likelihood of a sequence of hidden spikes $p_{\phi}(h_{\leq \mathcal{T}-1}\mid\mid v_{\leq \mathcal{T}-1})$. However, since data is observed without knowledge of the hidden spikes, the log-likelihood of the observed data is found by averaging over the hidden neurons activity as ${\mathcal{L}_{\upsilon_{\leq \mathcal{T}}}(\phi) = \log p_{\phi}(\upsilon_{\leq \mathcal{T}})} = \log\sum_{h_{\leq \mathcal{T}}} p_{\phi}(\upsilon_{\leq \mathcal{T}}, h_{\leq \mathcal{T}})$.
\begin{algorithm}[t]
\caption{OWOML-SNN}
\label{alg:owoml-snn}
\begin{algorithmic}[1]
\Require $\mathcal{B}^{(0)}$, hyperparameters $N, \mu$
\Repeat
    \State $t \gets t + 1$
    \State sample $T^{(t)} \in \mathcal{F}$
    \State initialize $D^{(t, 0)} \gets \emptyset$
    \While{data available for $T^{(t)}$}
        \State $i \gets i + 1$
        \State $D^{(t,i)} \gets D^{(t,i)} \cup z^{(t,i)}$
        \State $\phi^{(t,i)} \gets \Update\left( \theta^{(t,i)}, D^{(t,i)} \right)$
        \State inference using $\phi^{(t, i)}$
        \State sample meta-train data $\{D^{(n)}\}_{n=1}^N \in \mathcal{B}^{(t)}$
        \State $\theta^{(t, i+1)} \gets \MetaUpdate\left(\theta^{(t, i)}, \{D^{(n)}\}_{n=1}^N\right)$
    \EndWhile
    \State $\mathcal{B} = \mathcal{B} \cup D^{(t,i)}$
\Until{convergence}
\Statex
\Function{$\MetaUpdate$}{$\theta$, $\{D^{(n)}\}_{n=1}^N$}
    \For{$n = 1$ \textbf{to} $N$ parallel models}
        \State within-task training $\phi^{(n)} \gets \Update(\theta, D^{(n)})$
    \EndFor
    \State $\theta \gets \theta + \mu \sum_{n=1}^N \left( \theta - \phi^{(n)} \right)$
    \State \textbf{return} $\theta$
\EndFunction
\end{algorithmic}
\end{algorithm}

\section{Per-task Learning}\label{sec:withintask_update}

The per-task training function, $\Update(\phi, D)$ concatenates all the examples $(x_{\leq \mathcal{T}}, y_{\leq \mathcal{T}})\in D$ to obtain a single stream of data $(x_{\leq S}, y_{\leq S})$ with $S=M\mathcal{T}$. The function processes this data over time with the goal of addressing the maximum likelihood problem $\max_{\theta}\mathcal{L}_{\upsilon_{\leq S}}(\phi)$ with initialization $\theta$. Following \cite{jang2019introduction}, we approach this problem by maximizing the lower bound on the log-likelihood of the data 
\begin{align}\label{eq:vem_elbo}
    \mathcal{L}_{\upsilon_{\leq S}}(\phi) & \geq \mathrm{E}_{p_{\theta}(h_{\leq \!S\!-\!1}\!\mid\mid\! \upsilon_{\leq S\!-\!1})}\!\left[
    \log p_{\phi}(\upsilon_{\leq S}\!\mid\mid\! h_{\leq S-1})
    \right]\nonumber
    \\& \doteq  L_{\upsilon_{\leq S}}(\theta), 
\end{align} where the likelihood of the observed spikes $p_{\phi}(\upsilon_{\leq S}\mid\mid h_{\leq S-1})$ is defined as in (\ref{eq:glm_outprob}) with the spiking probability $\sigma(u_{i,\tau})$ generally dependent on the previous behavior of the hidden neurons.

The expected log-likelihood of the observed spikes (\ref{eq:vem_elbo}) is estimated via Monte Carlo sampling of the hidden spikes $[h_{i, \tau}\sim p_{\phi}(h_{i,\tau}\mid u_{i, \tau})]_{i\in\mathcal{H}}$ at every discrete time-instant $\tau$ which allows the computation of the new membrane potentials $[u_{i,\tau+1}]_{i\in\mathcal{V},\mathcal{H}}$ according to equation (\ref{eq:glm_mempot}). The SNN model parameter vector $\phi$, initialized by the hyperparameter $\theta$, is iteratively updated according to the gradient of (\ref{eq:vem_elbo}) calculated at every time $\tau$. The gradients with respect to the visible neuron model parameters can be derived directly from (\ref{eq:vem_elbo}) as \cite{jang2019introduction}
\begin{align}\label{eq:grad_local}
    & \nabla_{w^{\alpha}_{j,i}} \log p_{\theta_i}(\upsilon_{i,\tau}\mid u_{i,\tau}) = A^T\Vec{s}_{j,\tau-1}(\upsilon_{i,\tau}-\sigma(u_{i,\tau}))\nonumber \\
    & \nabla_{w^{\beta}_i} \log p_{\theta_i}(\upsilon_{i,\tau}\mid u_{i,\tau}) = B^T\Vec{s}_{i,\tau-1}(\upsilon_{i,\tau}-\sigma(u_{i,\tau}))\\
    & \nabla_{\gamma_i} \log p_{\theta_i}(\upsilon_{i,\tau}\mid u_{i,\tau}) = (\upsilon_{i,\tau}-\sigma(u_{i,\tau})). \nonumber
\end{align}
These derivatives include a post-synaptic error term $(\upsilon_{i,\tau} - \sigma(u_{i,\tau}))$ and a pre-synaptic term $A^T\Vec{s}_{j, \tau-1}$, where $\Vec{s}_{j,\tau-1} = [s_{j, \tau-1}, s_{j, \tau-2},...,s_{j,\tau-\tau_w}]^T$ is the $\tau_w \times 1$ window of pre-synaptic spikes that were processed at time $\tau$ and $A$ is the ${\tau_w\times K_a}$ matrix of basis vectors that define the pre-synaptic filter. For the hidden neurons, the online gradient is estimated by the REINFORCE gradient \cite{jang2019introduction}
\begin{equation}\label{eq:grad_hidden}
    \nabla_{\theta_{i\in\mathcal{H}}} L_{\upsilon_{\leq \mathcal{T}}}(\theta) \simeq \sum_{\tau=1}^{\mathcal{T}} \ell_{\tau} \nabla_{\theta} \log p(h_{\tau}\mid u_{\leq\tau-1})
\end{equation} where $\nabla_{\theta} \log p(h_{\tau}\mid u_{\leq\tau})$ is evaluated as in (\ref{eq:grad_local}), substituting $h_{i,\tau}$ for $\upsilon_{i,\tau}$, and we have the global error, or learning signal, 
\begin{equation}\label{eq:learning_sig}
    \ell_{\tau} = \sum_{i\in\mathcal{V}}\log\left(\overline{\upsilon}_{i,\tau}\overline{\sigma}(u_{i,\tau})+ \upsilon_{i,\tau}\sigma(u_{i,\tau}) \right).
\end{equation}  The derivatives in (\ref{eq:grad_local}) - (\ref{eq:grad_hidden}) are accumulated for $\Delta s$ discrete time-steps and added to an eligibility trace that is maintained as an exponentially decaying average of previous gradients to update the SNN model parameters. 

The gradients (\ref{eq:grad_local}) - (\ref{eq:grad_hidden}) give a three-factor update rule for the hidden neurons, where all computations are local aside from the global error signal. This can be interpreted as a signal as to how well the hidden neurons supports the generation of the desired output. The overall operation of the $\Update(\theta, D)$ function, including a sparsity-inducing regularizer \cite{jang2019introduction} is summarized in the supplementary materials.
\vspace{-0.1cm}
\section{SNN Meta-Learning}\label{sec:snn_ml}

We now describe OWOML-SNN, an online-within-online meta-learning algorithm for probabilistic SNNs that builds on the framework described in Sec. \ref{sec:owoml}. The overall algorithm is described in Algorithm \ref{alg:owoml-snn}, and is detailed next. The meta-learner is defined as an SNN model whose weights define the hyperparameters $\theta^{(t, i)}$. At each within-task time-step $(t, i)$, $N+1$ SNN models are instantiated with initial weight given by $\theta^{(t, i)}$. One SNN is used to carry out inference on the current task $T^{(t)}$, while the remaining $N$ SNN models are used to enable the meta-update. 

To elaborate, at every meta-time step $t$, a task $T^{(t)}\in\mathcal{F}$ is drawn, and the task-data buffer is initialized as $D^{(t,i)} = \emptyset$. Within-task data  is added to the current task-data buffer at every within-task time step $i$ in batches of $B$ training examples $z^{(t,i)} = \{(x^j_{\leq \mathcal{T}}, y^j_{\leq \mathcal{T}})\}_{j=1}^B$. The inference SNN implements online learning for the current task via the update function $\Update(\theta^{(t,i)}, D^{(t,i)})$ using the hyperparameter initialization $\theta^{(t, i)}$ and data in the task-data buffer.

To enable the meta update, the mentioned $N$ SNN models are trained online using $N$ data-sets sampled from the meta-data buffer $\{D^{(n)}\}_{n=1}^N \in \mathcal{B}^{(t)}$. Each of the data-sets includes $M$ training examples that are a subset of the data-set of a previously seen task such that ${D^{(n)} = \{(x^j_{\leq \mathcal{T}}, y^j_{\leq \mathcal{T}})\}_{j=1}^M}$. The hyperparameter is updated via the function $\MetaUpdate\left(\theta^{(t,i)}, \{D^{(n)}\}_{n=1}^N\right)$ which includes within-task training on the $N$ sampled data-sets. 

The inference accuracy of the within-task parameter $\phi^{(t, i)}$ is tested on a held out test data set for the current task. Online meta-learning is successful if the inference SNN is able to obtain satisfactory accuracy levels by using fewer examples for online adaptation. 

\begin{figure}[t]
    \centering
    \includegraphics[width=\linewidth, trim={0in, 0in, 0.5in, 0.5in}, clip]{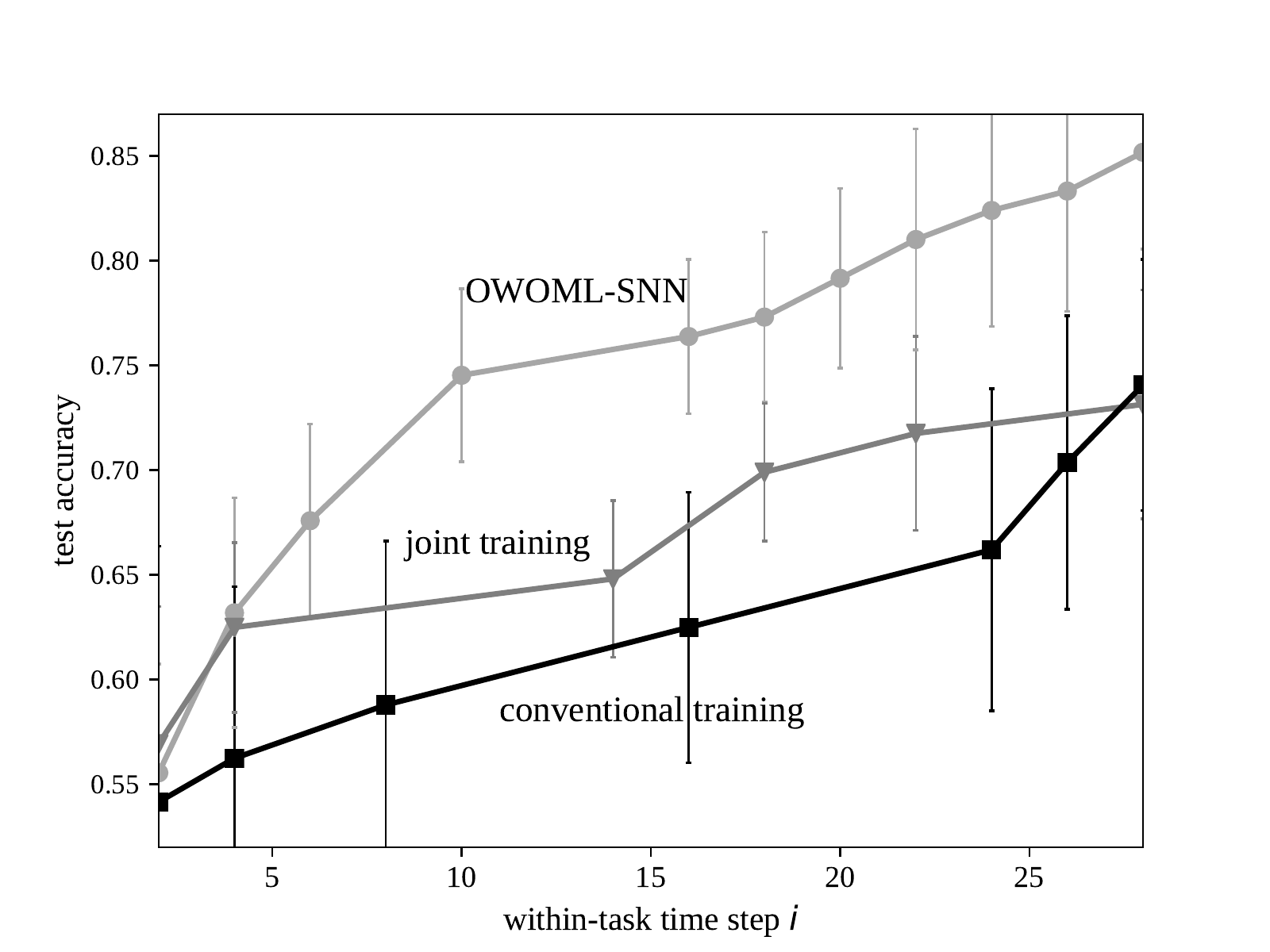}
    \caption{Test accuracy of within-task training after $t=15$ meta-time steps for the omniglot 2-way classification task family. Lines show an average over 6 new 2-digit classification tasks with half standard deviation error bars. 5-shot training is completed after $i=10$ within-task time steps}
    \label{fig:omniglotres}
    \vspace{-0.1cm}
\end{figure}
\begin{figure}
    \centering
    \includegraphics[width=\linewidth,trim={0in 0in 0.5in 0.5in}, clip]{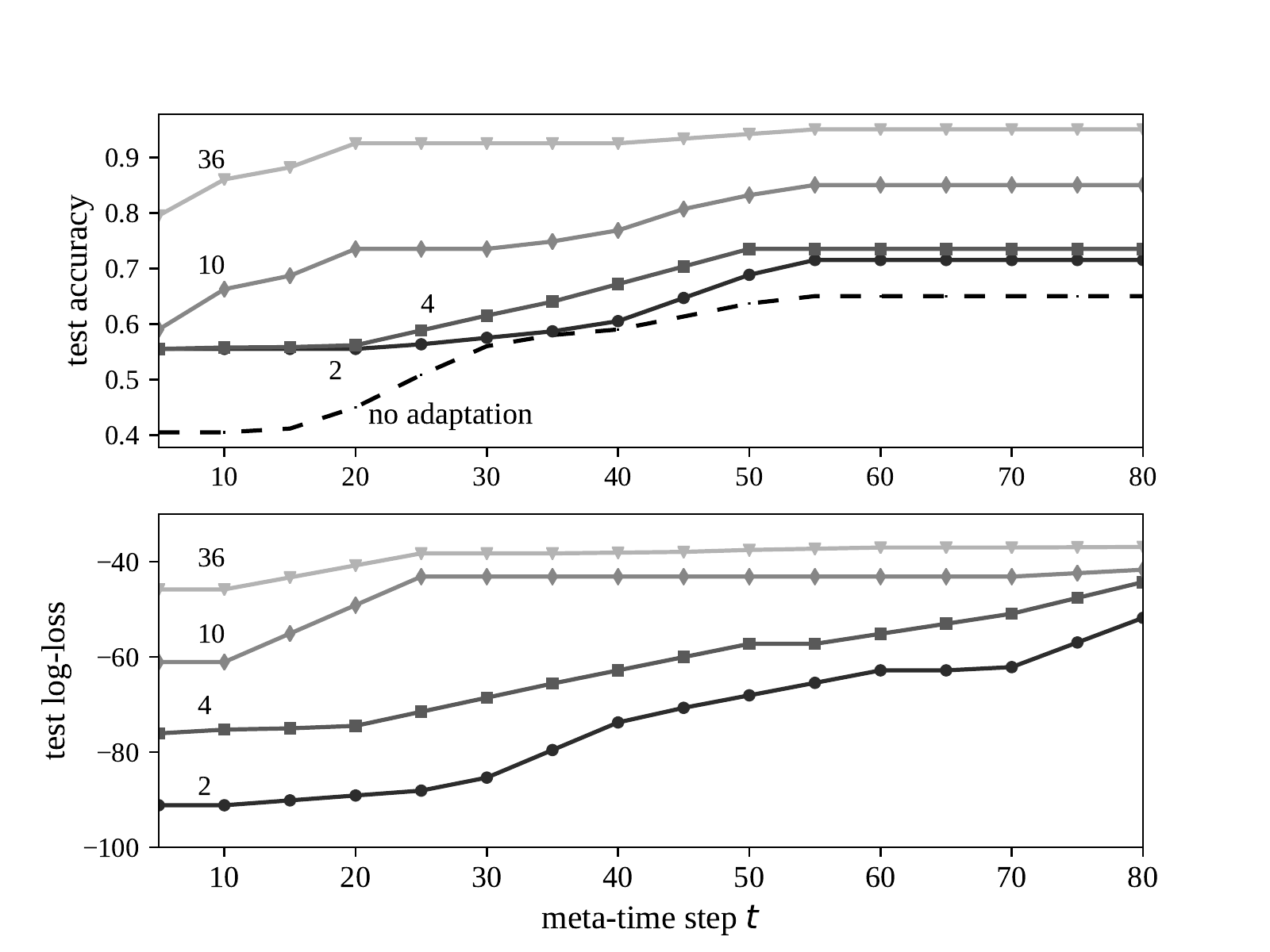}
    \caption{Within-task test accuracy (top) and loss (bottom) on the current task $T^{(t)}$ over online-within-online meta-training for the MNIST-DVS {2-way} classification task family after the full within-task dataset $D^{(t, i)}$ has been processed. The dashed line represents no within-task adaptation. Each solid line represents test results after within-task training on the number of training examples labeled. Due to variability in the hardness of the tasks, the lines show best accuracy seen so far, giving an envelope of the overall behavior.}
    \label{fig:accandloss_globaltraining}
\end{figure}
\section{Experiments}
\vspace{-0.1cm}
In this section, we compare the performance of OWOML-SNN to conventional per-task training and joint training (a standard benchmark for meta-learning \cite{zhou2018deep, amit2018meta, finn2017model}). Under conventional training, the meta-update function in Algorithm \ref{alg:owoml-snn} is disabled and the hyperparameter $\theta^{(t,0)}$ is randomly re-initialized for any new task $T^{(t)}$. Under joint training, the meta-training function $\MetaUpdate(\theta, \mathcal{B})$, is replaced by training across all tasks whose data is in buffer $\mathcal{B}^{(t)}$. This amounts to applying function $\Update(\theta^{(t, i)}, D)$, where $D$ includes examples sampled from $\mathcal{B}^{(t)}$ as for the meta-update function.

We first consider the family of omniglot 2-way classification tasks \cite{lake2015human}, which is often used to test the within-task generalization capabilities of meta-learners \cite{finn2017model, scherr2020one, vinyals2016matching, koch2015siamese}. We follow the more complex definition of the problem in which the two characters to be classified in each task may be from different alphabets. Each task dataset includes 14 examples from each class, while 6 examples from each class are reserved for the test dataset. The figures are downsized to $26\times 26$ pixels and rate encoded to obtain examples of the form $(x_{\leq \mathcal{T}}, y_{\leq \mathcal{T}})$ where $y_{\leq \mathcal{T}}$ is a one hot encoded label and $\mathcal{T}=80$ \cite{jang2019introduction}. 

We train a fully connected GLM-based SNN with 4 hidden neurons and 2 visible neurons with additional lateral connections between the hidden neurons and feedback connections from the visible neurons to the hidden neurons. After 15 meta-time steps, i.e., at $t=15$, we test the hyperparameter initialization on the next 6 unseen tasks without further meta training updates. We set $M=2, N=5$, and $\Delta s = 5$. The results in Fig. \ref{fig:omniglotres} show that after training on 5 examples from each class,  OWOML-SNN provides an increase in accuracy over the baseline of conventional training that is about $18\%$ larger than joint training. After further training, we observe that OWOML-SNN enables the SNN to achieve on average a $15\%$ higher accuracy overall than the two benchmarks.

We now show that the fast adaptation capability of OWOML-SNN extends to the case of a continuous input stream with data encoded in the spike timing, by considering 2-way classification on the the MNIST-DVS data-set \cite{serrano2015poker}. MNIST-DVS is a dataset of MNIST images captured by a neuromorphic DVS camera that generates localized events based on changes in individual pixels over time. The tasks are sampled from the set of permutations of the digits between 0 and 6 with 900 examples from each class used for training and 100 held out for test. Each image is cropped to the $26 \times 26$ pixel size and the events are downsampled to create a sequence of length $\mathcal{T} = 50$ \cite{skatchkovsky2020federated}.

 We train an SNN with the same recurrent architecture as described in the previous experiment with hyperparameters $M=4, N=5$, and $\Delta s = 20$. We examine the performance of the SNN instantiated for task-specific training on the current task $T^{(t)}$ as the hyperparameter initialization is improved over meta-training. We show results for online adaptation after the full dataset $D^{(t, i)}$ for the current task has been processed i.e., at the maximum value of within-task time step $i$. The results in Fig. \ref{fig:accandloss_globaltraining} confirm that OWOML-SNN enables fast online adaptation that continuously improves as the number of meta-time steps $t$ increases and more meta-updates are applied, requiring fewer examples to achieve within-task accuracy targets.

\section{Conclusions}
We have presented OWOML-SNN, an online-within-online meta-learning algorithm that builds on online variational learning algorithm for probabilistic SNNs. We have demonstrated the performance benefits of OWOML-SNN when adapting to new tasks observed sequentially in an online fashion. OWOML-SNN is based on a first-order approximation of the meta-gradient, yielding a local meta-learning rule that follows the principles of predictive coding, with no reliance on back-propagation of gradients.  This makes the approach a prime candidate for fast on-chip adaptation to new tasks, and in particular for tasks that involve streaming input data. While the first order approximation yields an efficient local update rule, it also generally reduces the capacity of adaptation to new tasks \cite{nichol2018first}. As future work, it would be interesting to investigate more accurate approximations of the meta-gradient that retain the property of locality, as well as convergence properties. 




\bibliographystyle{IEEEtran}
\bibliography{IEEEabrv, references}



\end{document}


\title{Supplementary Materials for \\Fast On-Device Adaptation for Spiking Neural Networks via Online-Within-Online Meta-Learning}
\author{Anonymous DSLW Submission}
\maketitle
\section{More Discussion About Related Work}
Many of the meta-learning algorithms for ANN models reviewed adopt local learning rules in the inner loop \cite{gu2019meta,metz2018meta,najarro2020meta} or during adaptation to a new task \cite{nguyen2018meta,miconi2018differentiable} as inspired by biological learning mechanisms. Similarly, SNNs are being explored for meta-learning with the dual focus of biological plausibility and low-power neuromorphic on-chip learning. The outer loop of training in all reviewed cases is implemented offline, with updates in the form of surrogate gradient back-propagated errors averaged over training examples from many tasks \cite{scherr2020one, stewart2020chip, bellec2018long} or a zeroth order meta-gradient that evaluates error as a Function of the inner loop performance \cite{bohnstingl2019neuromorphic}. With the goal of enabling local learning on neuromorphic chips, inner-loop updates are online and local except \cite{bellec2018long}. The outer loop of training is used to enable the locality of inner loop updates by learning a parameterized learning rule \cite{bohnstingl2019neuromorphic, scherr2020one} or by pre-training a fixed portion of the model and learning only a read-out layer \cite{stewart2020chip,stewart2020online}.

The spiking models discussed require surrogate gradients for BP due to the use of the deterministic leaky integrate and fire (LIF) neuron model. Instead we adopt a probabilistic neuron model that supports maximum likelihood learning thereby avoiding the issues of credit assignment over time and the non-differentiable spike activation Function. Several probabilistic neuron models were introduced and studied in the context of computational neuroscience such as bayesian spiking neurons (BSN) \cite{deneve2008bayesian, nessler2013bayesian}, the stochastic spike response model (SRM) \cite{buesing2011neural} and the generalized linear model (GLM) \cite{pillow2008spatio}. BSNs and stochastic SRMs have been used for machine learning with the model being trained either via unsupervised STDP \cite{kuhlmann2014approximate, tavanaei2015studying, al2015inherently, pyle2019subthreshold} or top-down gradient methods \cite{hu2018efficient}. Aside from the biological plausibility of probabilistic synapses \cite{llera2019computational} it has recently been proposed that synaptic plasticity in the brain implements some form of probabilistic learning \cite{zheng2020probabilistic}. Probabilistic learning methods include unsupervised learning \cite{zhang2019information}, contrastive divergence to estimate maximum likelihood learning for LIF neurons \cite{neftci2016stochastic} and stochastic sampling to optimize for sparse network connectivity where the parameters are learned via BPTT \cite{bellec2017deep}. 
\vfill \null
\section{Per-Task Learning -- Algorithmic Table}
\begin{algorithm}[H]
\caption{$\Update(\theta, D)$}
\label{alg:update}
\begin{algorithmic}[1]
\Require $\theta, D$, hyperparameters $\kappa, \Delta s, \eta $
\State initialize $\phi \gets \theta$, $l_{0}\gets 0$,  $e_{i,0} \gets 0$
\For{$\tau=1$ \textbf{to} $S$}
    \State compute membrane potentials $u_{i,\tau}$  as in (Eq. ($4$)) 
    \For{$i\in\mathcal{H}$}
        \State Monte Carlo sampling of $h_{i,\tau}$
    \EndFor
    \State accumulate learning signal $l_{\tau} = l_{\tau-1} + \ell_{\tau}$ (Eq. ($10$))
    \State accumulate local gradients
    
    \begin{center} ${e_{i, \tau} =  e_{i, \tau-1} + \nabla_{w^{\alpha}_{j,i}}\log p(\upsilon_{i, \tau}\mid\mid h_{\leq\tau-1})}$ (Eq. ($8$))\end{center}
\If{$\tau$ is a multiple of $\Delta s$}
    \State compute learning signal trace\\ 
    \begin{equation}\label{eq:trace_learningsig}
        l = \kappa l + (1-\kappa) l_{\tau}
    \end{equation}
    \For{neurons $i\in\{\mathcal{V},\mathcal{H}\}$}
        \State compute local gradient traces 
        \begin{align}\label{eq:trace_localgrad}
        & e_i = \kappa e_i + (1-\kappa) e_{i,\tau}
        \end{align}
        \State update parameters

        \begin{center}
            $\phi_i \gets \phi_i + \eta \begin{cases}
                    e_i \text{ if } i\in \mathcal{V}\\
                    l e_i \text{ if } i \in \mathcal{H}
            \end{cases}$
        \end{center}
    \EndFor
    \State reset $l_{\tau} \gets 0$, $e_{i,\tau}\gets0$ 
\EndIf
\EndFor
\State \textbf{return:} $\phi$
\end{algorithmic}
\end{algorithm}
\bibliographystyle{IEEEtran}
\bibliography{IEEEabrv, references}